\documentclass[letterpaper]{article} 
\usepackage{arxiv}  
\usepackage[hyphens]{url}  
\usepackage{graphicx} 
\usepackage{natbib}  
\usepackage{caption} 
\usepackage{booktabs}

\usepackage{amsmath}
\usepackage{amssymb}

\usepackage{multirow}

\usepackage{array}
\usepackage{listings}
\title{MedTextWeaver: Procedural Knowledge Evolution in Agentic Medical Text Editing}

\author{
    Ziyan Xiao, Yinghao Zhu, Liang Peng, Kyongtae T Bae, Lequan Yu
}

\affiliations{
    The University of Hong Kong
}

\begin{document}
\maketitle

\begin{abstract}
Medical text editing is essential for improving communication among diverse stakeholders in clinical settings. However, adapting LLM agents to this task remains challenging because expert supervision is often sparse, fragmented, and distributed across interacting quality dimensions. We identify that direct accumulation or retrieval of individual feedback is insufficient for effective adaptation, as fragmented evaluations do not directly translate into a coherent understanding of medical text quality.
Based on this observation, we propose \textbf{MedTextWeaver}, a training-free framework that transforms fragmented evaluative evidence into global quality principles and actionable procedural knowledge for medical text editing. Across three clinical text datasets and a real-world validation experiment, MedTextWeaver consistently improves performance over strong LLM baselines and existing memory-based adaptation approaches. Further analysis demonstrates that the learned knowledge enables more effective adaptation under limited supervision while providing an explicit and interpretable interface between expert evaluations and LLM editing behavior.
\end{abstract}

\begin{figure}[t]
  \centering
  \includegraphics[width=\columnwidth]{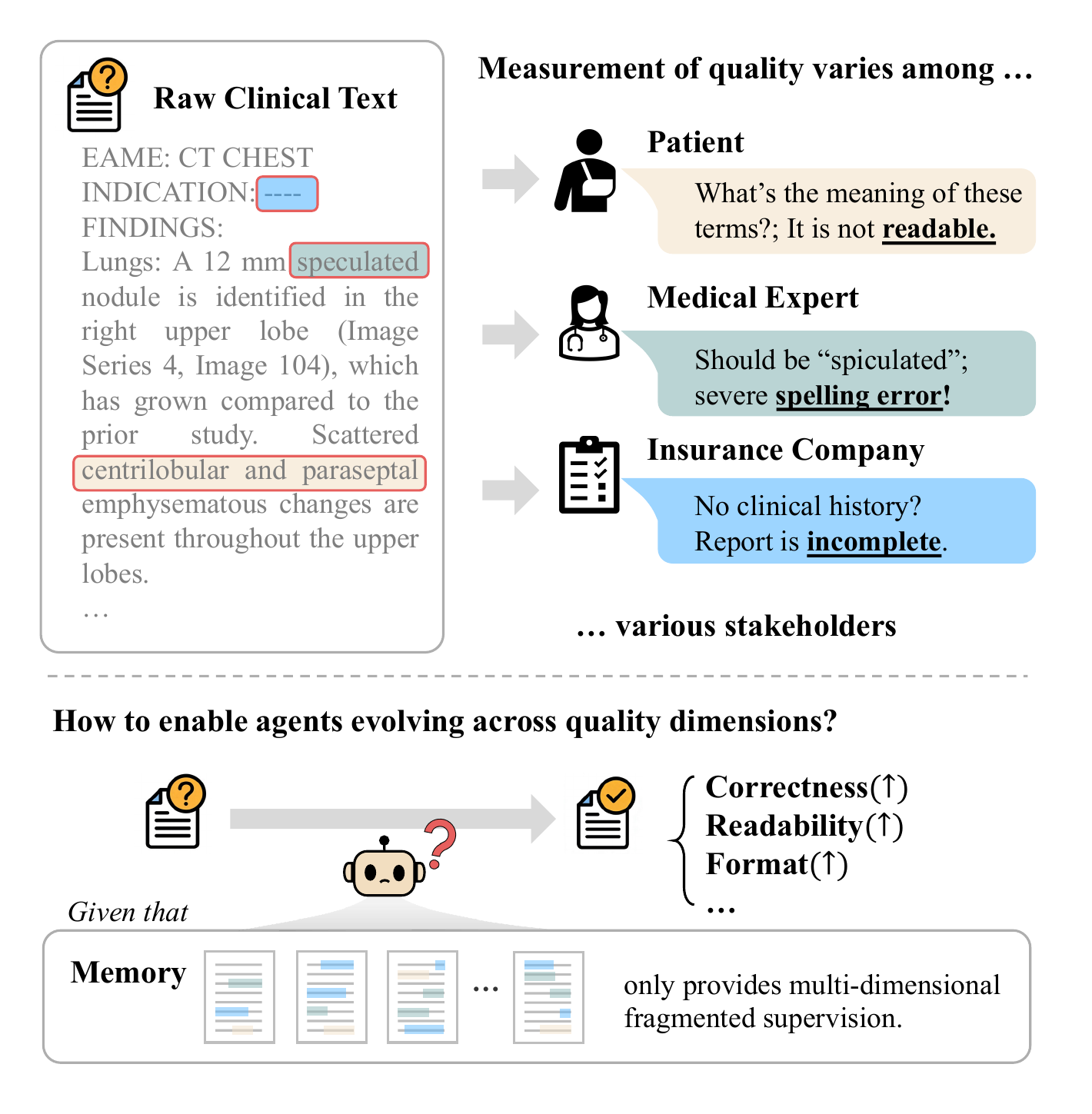}
  \caption{Project Motivation. Medical text editing agents suffer from challenges of small-sample, fragmented, and multi-dimensional supervision provided by various stakeholders.}
  \label{fig:problem_begins}
\end{figure}

\begin{figure*}[t]
  \centering
  \includegraphics[width=\textwidth]{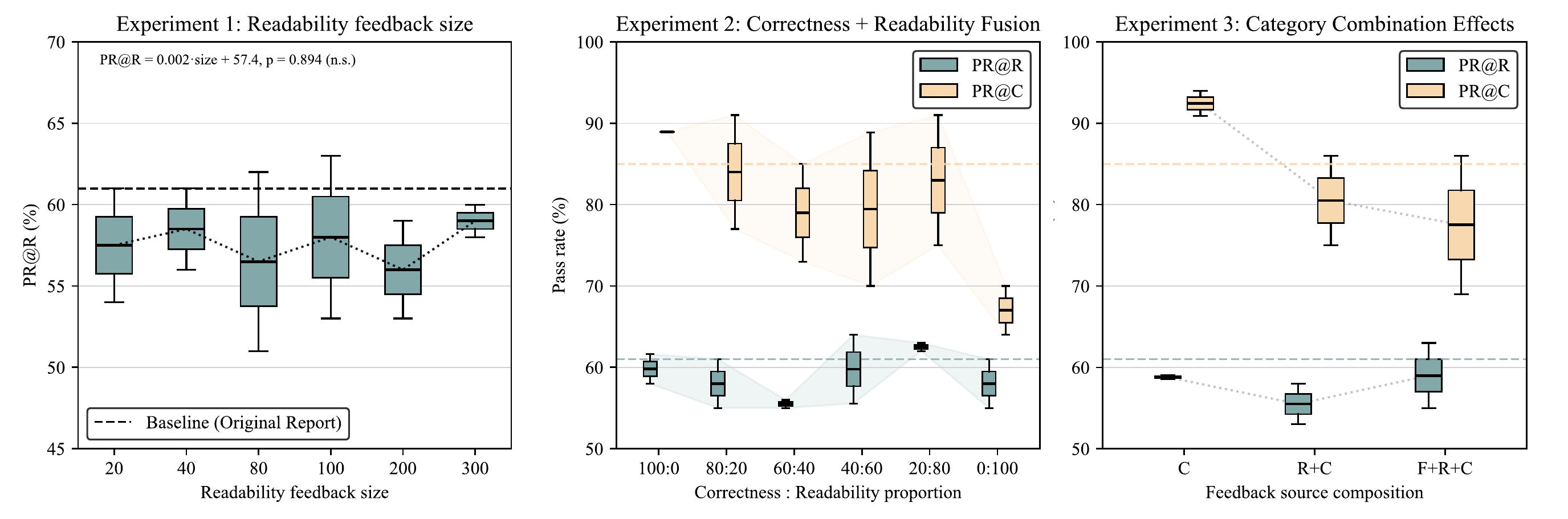}
  \caption{Preliminary Experiment. (a) RAG-based methods in medical text editing fail to exhibit linear scaling as feedback accumulates. (b) Fusion of multiple feedback sources leads to large variance in pass rates, indicating unsettled confusion brought by heterogeneous feedback. (c) Performance decreases when feedback is accumulated solely across quality dimensions, even assuming the same amount of feedback per quality dimension.}
  \label{fig:preliminary_exp}
\end{figure*}

\section{Introduction}
\label{sec:introduction}

The clinical domain generates large volumes of informative but imperfectly organized text. Processing such data requires substantial professional expertise and adaptation to diverse communication needs, while the associated documentation burden remains a persistent challenge in healthcare systems~\cite{arndt2017tethered,sinsky2016allocation,tai2017electronic}. Consequently, \textbf{medical text editing}, which improves various quality dimensions of clinical documents, is increasingly important for efficient and reliable healthcare communication~\cite{perkins2024improving,bergomi2024reshaping}. Compared to general text editing, clinical text editing requires satisfying more quality requirements simultaneously, such as medical information correctness, format compliance, and readability for various stakeholders.

These characteristics create a distinctive adaptation problem for large language models (LLMs). First, \textbf{evaluative supervision is fragmented}: feedback is often provided as local judgments about individual errors, specific text spans, or particular quality dimensions rather than a complete specification of desired behavior. Such sparse evidence provides useful but incomplete information for correcting a model's systematic biases. Second, \textbf{quality dimensions interact}: improvements in one aspect of a document may affect other aspects, such that optimizing individual dimensions independently does not necessarily produce globally improved text. The resulting supervision is therefore not only limited in quantity, but also distributed across partial and potentially interdependent evaluations.

While LLMs have demonstrated substantial potential for medical assistance~\cite{sandmann2025benchmark,moor2023foundation}, this supervision regime creates a fundamental challenge for medical text editing. The scarcity and fine-grained nature of expert feedback make repeated supervised fine-tuning or reinforcement learning costly and difficult to scale~\cite{ouyang2022training}. Training-free approaches, including in-context learning, retrieval, and memory-based adaptation, avoid parameter updates but typically reuse individual examples, evaluations, or compressed summaries of previous observations~\cite{lewis2020retrieval}. These approaches show limited adaptation capabilities because individual examples lack representativeness and reusable guidance is insufficient for adaptation. We further illustrate the limitations through preliminary experiments in Section~\ref{sec:diagnosing_adaptation_failure}. The central challenge is therefore not simply how to access or store scarce feedback, but \textbf{how to transform fragmented evaluative evidence into knowledge that can reshape model behavior.}


Through our investigation, we identify that effective adaptation requires transforming heterogeneous evaluations into coherent and actionable knowledge of quality requirements. Based on this insight, we develop \textbf{MedTextWeaver}, a training-free framework for procedural knowledge evolution in agentic medical text editing. MedTextWeaver progressively reorganizes evaluative evidence through hierarchical knowledge distillation, cross-dimensional synthesis, and procedural knowledge formation. 
Experiments across three clinical text editing datasets demonstrate that MedTextWeaver consistently outperforms strong LLM baselines and existing training-free adaptation methods under limited supervision. Further real-world validation shows that the learned knowledge improves medical error detection, while case studies reveal the interpretability of the editing principles distilled from distributed expert evaluations. These results suggest that scalable adaptation under scarce clinical supervision depends not only on accessing and accumulating feedback, but also on evolving fragmented evidence into globally coherent and reusable knowledge.


Our contributions are threefold.

\begin{itemize}
    \item We characterize the adaptation challenges of medical text editing under fragmented and multi-dimensional clinical supervision, showing that feedback accumulation alone does not guarantee effective behavioral adaptation.
    \item We propose MedTextWeaver, a training-free knowledge evolution framework that transforms sparse feedback into cross-dimensional and reusable editing knowledge.
    \item We demonstrate that structured knowledge transformation enables efficient adaptation under limited supervision, improves real-world medical error detection, and provides an interpretable interface for examining learned editing behaviors.
\end{itemize}

\begin{figure*}[t]
  \centering
  \includegraphics[width=\textwidth]{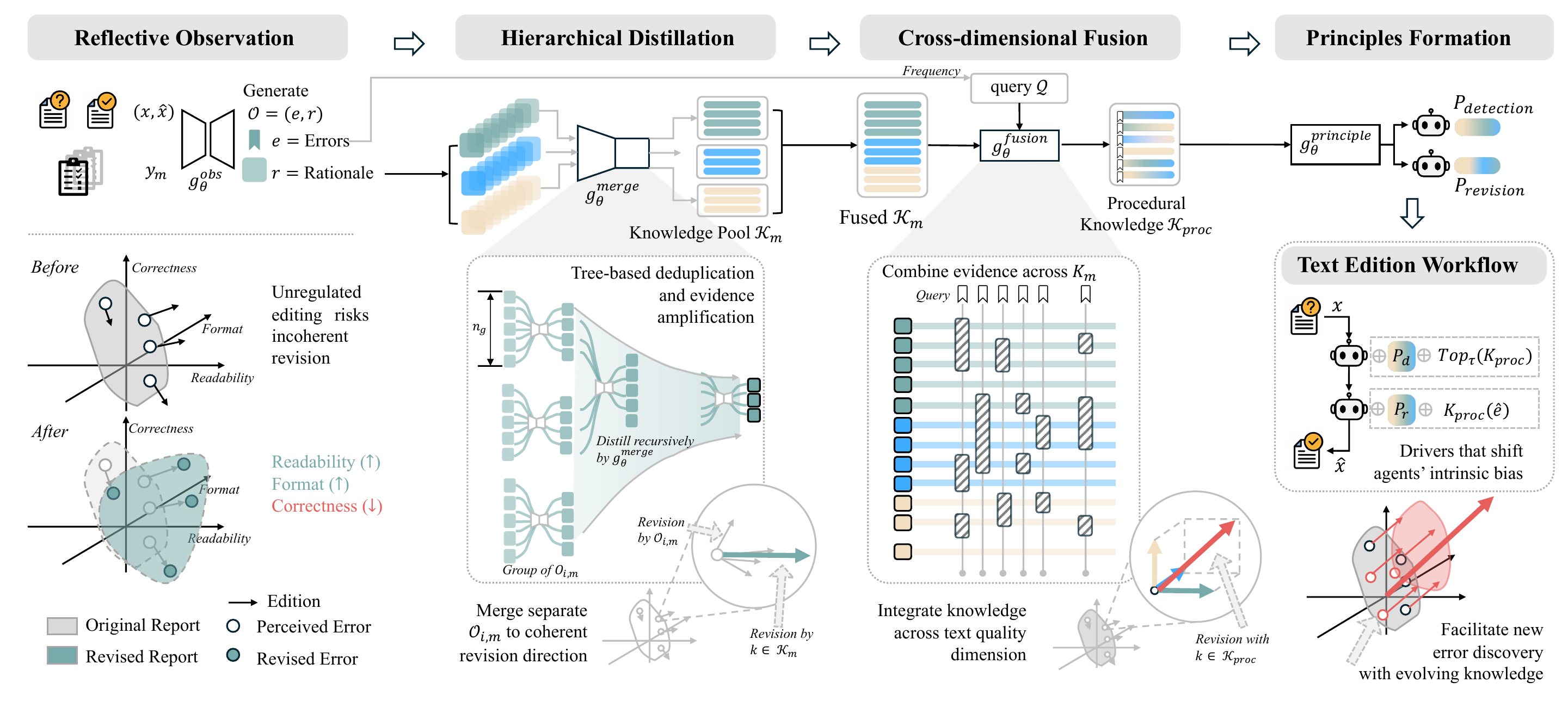}
  \caption{The proposed MedTextWeaver pipeline. It demonstrates the process of forming knowledge from raw feedback to distilled multi-layered experience injected into the two phases of the agentic pipeline: error detection and text revision.}
  \label{fig:pipeline}
\end{figure*}

\section{Diagnosing Adaptation Failure in Medical Text Editing Agents}
\label{sec:diagnosing_adaptation_failure}

We first investigate how LLM agents adapt when supervision is provided as sparse and fragmented evaluations of medical text quality. We collected 100 chest X-ray radiology reports and 100 discharge summaries from the MIMIC-IV datasets. Following the evaluation settings in Section~\ref{sec:experimental_setups}, we considered two representative quality dimensions, correctness and readability. Figure~\ref{fig:preliminary_exp} summarizes three preliminary analyses examining whether evaluative evidence can be effectively converted into improved editing behavior.

Specifically, we investigate three questions: (1) whether increasing feedback quantity leads to proportional adaptation gains; (2) whether different quality dimensions can be optimized independently; and (3) whether accumulating evaluations across dimensions naturally leads to improved editing quality.

\subsection{Feedback Quantity Does Not Determine Adaptive Gain}
\label{subsec:diagnosing_feedback_quantity}

We first examine whether increasing the amount of evaluative supervision directly improves editing performance. We progressively increase the number of readability-related feedback records and measure the pass rate of edited texts. As shown in Figure~\ref{fig:preliminary_exp}(a), additional feedback provides limited improvement, indicating that the amount of available supervision alone does not determine adaptation effectiveness. This observation suggests that evaluative evidence cannot be effectively utilized through direct accumulation, motivating mechanisms that transform fragmented feedback into more informative representations.

\subsection{Interacting Quality Dimensions Produce Imbalanced Adaptation}
\label{subsec:diagnosing_interacting_dimensions}

We next examine whether individual quality dimensions can be optimized independently. Using a fixed set of 200 feedback records, we vary the proportion of correctness and readability supervision and evaluate the resulting changes across both dimensions. As shown in Figure~\ref{fig:preliminary_exp}(b), incorporating feedback from multiple dimensions increases variance in LLM performance, while the upper bounds of achievable performance remain relatively unaffected. This indicates substantial interaction effects between quality objectives and points to a promising space for quality evolution in multiple dimensions. Effective adaptation therefore requires integrating knowledge across multiple evaluation perspectives to balance improvements and understand the relationship between them.

\subsection{Accumulated Evaluations Do Not Guarantee Coherent Adaptation}
\label{subsec:diagnosing_accumulated_evaluations}

We further investigate whether continuously accumulating evaluations from multiple sources yields consistent improvements across quality dimensions. Figure~\ref{fig:preliminary_exp}(c) shows that additional evaluations do not necessarily translate into further gains and may even introduce degradation in certain dimensions. Importantly, this phenomenon is non-uniform: feedback from one dimension can contain transferable information that benefits another. These findings highlight the limitations of unstructured accumulation and motivate the need for organizing distributed evaluations into higher-level representations that capture shared behavioral patterns, thereby enabling coherent adaptation across dimensions.



\section{Method}
\label{sec:method}

We introduce MedTextWeaver, a training-free framework that transforms fragmented evaluative supervision into structured knowledge for medical text editing. 

The framework consists of four stages:
(1) reflective observation construction, which converts score evaluations into textual evidence;
(2) hierarchical knowledge distillation, which aggregates fragmented observations into robust revision knowledge;
(3) cross-dimensional knowledge synthesis, which integrates interacting quality requirements; and
(4) knowledge-guided editing, which activates the learned knowledge during agent execution.

Throughout the framework, we use a frozen LLM parameterized by $\theta$. The LLM configurations are provided in the supplementary material. 
Different functions $g_{\theta}^{(\cdot)}$ denote different prompting strategies applied to the same underlying model, where the prompts for each $g_{\theta}^{(\cdot)}$ are provided in the supplementary material.

\subsection{Problem Formulation}
\label{subsec:method_problem_formulation}

Given a collection of medical documents $\mathcal{X}=\{x_i\}_{i=1}^{N}$, each document is improved by an editing agent to obtain $\hat{x}_i$. 
External evaluators provide feedback from multiple quality dimensions $\mathcal{M}$, such as correctness, readability, and format compliance.

We represent the evaluation for document $i$ under dimension $m$ as:

\[
O_{i,m}=g_{\theta}^{obs}(x_i,\hat{x}_i,y_{i,m}),
\]

where $y_{i,m}$ denotes the evaluation score and $O_{i,m}=(e_{i,m},r_{i,m})$ contains the identified issue and corresponding rationale.

Conventional retrieval-based adaptation treats these observations as independent examples and aims to retrieve similar historical cases. However, preliminary analysis shows that fragmented evaluations provide only local evidence of multiple quality requirements. We therefore aim to construct a knowledge representation $\mathcal{K}$ that captures the shared behavioral patterns underlying diverse evaluations. 

\subsection{Hierarchical Knowledge Distillation}
\label{subsec:method_hierarchical_distillation}

Individual evaluations are often highly specific and noisy. This poses challenges for memory usage: separating noise from general error types and transferring local revision patterns to global reusable rules.

To address these issues, MedTextWeaver performs hierarchical knowledge distillation. The observations are recursively merged:

\[
V^{(t+1)}
=
g_{\theta}^{merge}
(\{v_j^{(t)}\}_{j=1}^{n_g}),
\]

where $n_g$ observations are combined at each iteration.

The hierarchical process serves two purposes.
First, it consolidates repeated patterns across evaluations and reduces redundancy caused by isolated examples.
Second, complementary revision suggestions are combined, allowing the model to identify broader editing requirements rather than individual mistakes.

After recursive synthesis, the final knowledge pool is formed by disjoint rules, $\mathcal{K}_m=\{k_1,k_2,...,k_L\}, \quad m\in\mathcal{M}$. Each knowledge unit represents a generalized relationship between an error pattern and its corresponding improvement direction. These steps not only reduce the burden of retrieval, but also amplify the revision capability of each knowledge pool.

\subsection{Cross-Dimensional Knowledge Synthesis}
\label{subsec:method_cross_dimensional_synthesis}

Although hierarchical distillation improves generalization within each evaluation dimension, effective medical text editing requires satisfying multiple interacting quality dimensions simultaneously. Therefore, we introduce cross-dimensional synthesis by reorganizing knowledge according to prevalent error types revealed in the original reflections in Section~\ref{subsec:method_problem_formulation}.

We first identify knowledge queries from dominant error patterns:

\[
\mathcal{Q}
=
Freq(\{e_{i,m}\}).
\]

For each dominant error type $q$, knowledge from different quality dimensions is integrated:

\[
K_q
=
g_{\theta}^{fusion}
(
q,
\{k_m\}_{m\in\mathcal{M}}
),
\]

where $k_m$ denotes knowledge derived from quality dimension $m$ in Section~\ref{subsec:method_hierarchical_distillation}.

This operation changes the organization principle from dimension-specific optimization to problem-oriented synthesis. The final procedural knowledge pool is $\mathcal{K}_{proc}=\{K_q|q\in\mathcal{Q}\}$.

\begin{figure}[t]
  \centering
  \includegraphics[width=\columnwidth]{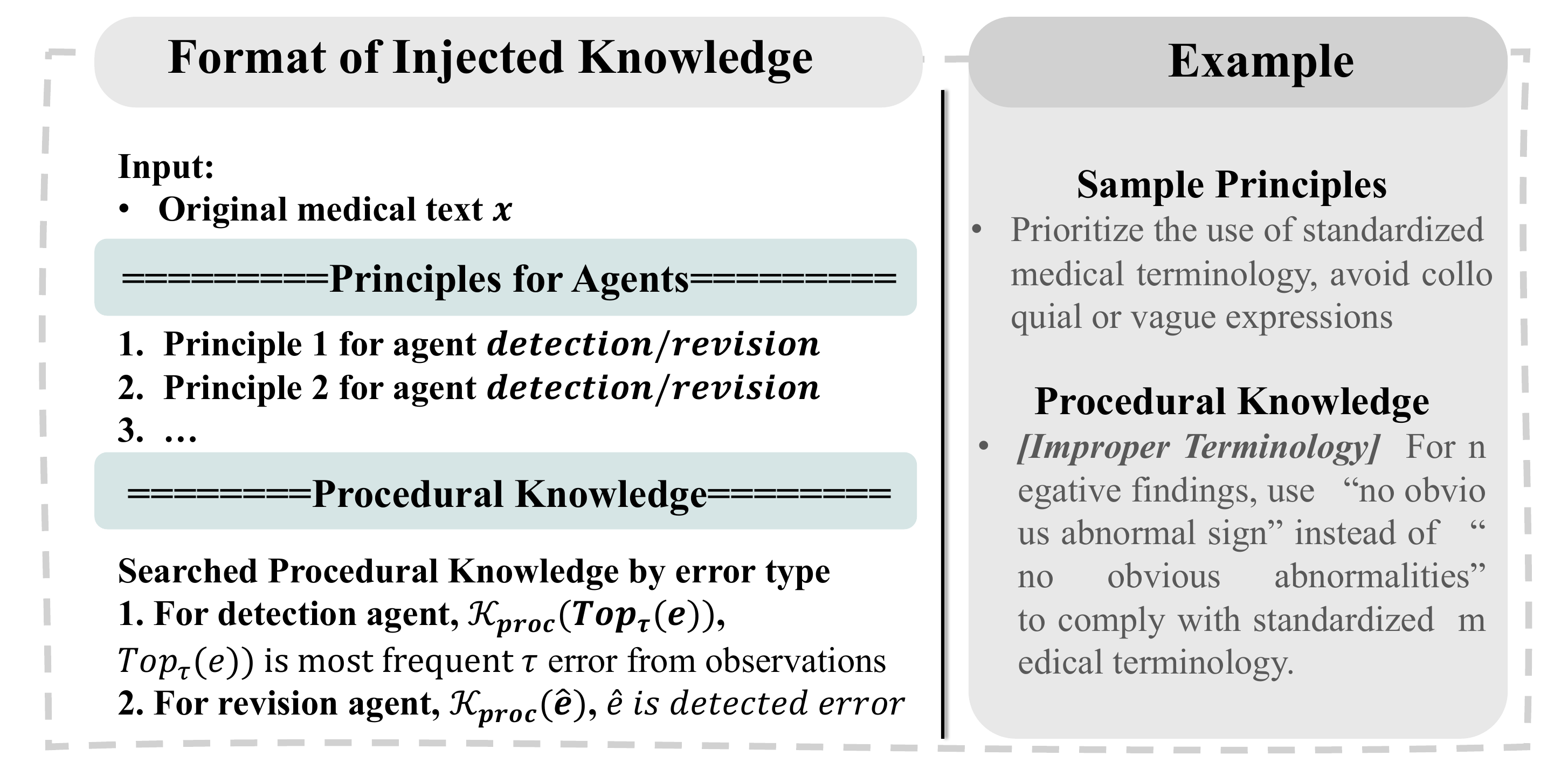}
  \caption{Example of knowledge injected into the detection and revision agents.}
  \label{fig:example_context}
\end{figure}

\subsection{Principle Formation and Agent Workflow}
\label{subsec:method_procedural_formation}

Procedural knowledge provides actionable guidance for frequently observed errors but remains limited to known error patterns. To identify broader quality requirements beyond individual cases, MedTextWeaver further abstracts high-level editing principles.

Specifically,

\[
P_a
=
g_{\theta}^{principle}
(\mathcal{K}_{proc}, a),
\]

where $P$ represents generalized principles describing desirable editing behaviors for agent $a \in \{detection, revision\}$.

Finally, we validate the synthesized knowledge within the editing workflow.
To simplify the problem, we choose two essential agents, an error detection agent and a text revision agent, to form an execution environment for medical text editing.

For an input document $x$, the detection stage receives:

\[
\hat{e}
=
g_{\theta}^{detect}
(
x,
P,
Top_{\tau}(\mathcal{K}_{proc})
),
\]

where $Top_{\tau}$ selects the most relevant procedural knowledge items to control context length.

After identifying potential errors, the revision stage applies corresponding knowledge:

\[
\hat{x}
=
g_{\theta}^{revise}
(
x,
P,
K_{\hat{e}}
).
\]

Thus, the same knowledge representation supports both global error awareness through principles and detailed correction through procedural knowledge. Figure~\ref{fig:example_context} illustrates the context injected into the LLM in MedTextWeaver. Implementation details of MedTextWeaver, including prompting templates, knowledge examples, and retrieval settings, are provided in the supplementary material.

\begin{table*}[t]
\centering
\small
\setlength{\tabcolsep}{1mm}
\begin{tabular}{lcccccccccccc}
\hline
Data Source & \multicolumn{4}{c}{MIMIC Chest X-ray} & \multicolumn{4}{c}{MIMIC Discharge} & \multicolumn{4}{c}{In-house Abdominal CT}  \\
\cline{2-13}
Pass Rate & PR@C & PR@R & PR@F & OPR & PR@C & PR@R & PR@F & OPR & PR@C & PR@R & PR@F & OPR \\
\hline
\quad Original Report & 100.0 & 54.50 & 17.09 & 9.55 & 100.0 & 39.25 & 41.40 & 15.59 & 100.0 & 62.00 & 30.50 & 19.00 \\
\multicolumn{13}{l}{\textbf{Section 1: SOTA LLMs Baseline}} \\
\quad Deepseek-v4-pro & 85.13 & 49.23 & 17.44 & 7.18 & 74.53 & 46.58 & 40.37 & 12.42 & 99.00 & 62.50 & 51.00 & 33.50 \\
\quad Gemini-3.5-flash & 75.00 & 49.00 & 61.50 & 26.00 & 96.50 & \textbf{46.00} & 42.50 & 15.00 & 96.00 & 62.00 & 21.50 & 15.00 \\
\multicolumn{13}{l}{\textbf{Section 2: Comparative RAG/Memory Methods}} \\
\quad RAG (Detection) & 84.00 & 53.50 & 20.00 & 9.50 & 97.99 & 39.70 & 46.37 & 16.08 & 92.00 & 58.50 & 41.00 & 25.00 \\
\quad RAG (Revision) & 84.00 & 56.00 & 16.50 & 7.50 & 97.97 & 36.04 & 41.12 & 11.17 & 97.00 & 59.00 & 47.50 & 30.00 \\
\quad AMW & 87.50 & 55.96 & 16.58 & 8.29 & 93.00 & 32.00 & 44.00 & 11.00 & 100.0 & 58.00 & 45.00 & 31.00 \\
\quad MemoryBank & 91.33 & 56.63 & 46.00 & 11.50 & 90.50 & 31.50 & 46.00 & 11.50 & 99.00 & 59.00 & 42.00 & 26.50 \\
\quad ExpeL & 91.33 & 56.63 & 16.33 & 8.16 & 94.50 & 33.00 & 2.00 & 0.00 & 99.50 & 59.50 & 17.00 & 10.00 \\
\quad EvolveR & 85.00 & 56.00 & 28.00 & 18.00 & 95.48 & 35.18 & \textbf{49.75} & 15.07 & 99.50 & 58.50 & 43.00 & 28.00 \\
\multicolumn{13}{l}{\textbf{Section 3: Weaving Experience}} \\
\quad Deepseek-v4-flash & 74.50 & 56.50 & 21.00 & 10.00 & 94.50 & 37.00 & 47.00 & 15.50 & \textbf{99.50} & 59.00 & 28.00 & 18.00  \\
\quad + MedTexWeaver(Detection) & \textbf{98.49} & 55.28 & 75.88 & 43.72 & 99.50 & 43.81 & 43.30 & 19.07 & 86.00 & \textbf{63.00} & 52.00 & 29.50  \\
\quad + MedTexWeaver(Revision) & 84.90 & 55.73 & 63.02 & 31.77 & 99.50 & 41.33 & 46.43 & 18.37 & 99.00  & 59.50 & 51.50 & 34.50\\
\quad + MedTexWeaver(Full) & 95.43 & \textbf{58.38} & \textbf{77.66} & \textbf{44.67} & \textbf{99.50} & 43.37 & 48.47 & \textbf{19.90} & 88.00 & 62.50 & \textbf{57.00} & \textbf{35.00}  \\
\hline
\end{tabular}
\caption{
Comparative experimental results across data sources and evaluation metrics. PR@C = pass rate for correctness, PR@R = pass rate for readability, PR@F = pass rate for format, and OPR = overall pass rate when all three metrics pass.
}
\label{tab:main}
\end{table*}

\section{Experimental Setups}
\label{sec:experimental_setups}

\subsection{Datasets}
\label{subsec:experimental_datasets}
We used three clinical text datasets in English and Chinese: (1) \textbf{MIMIC Radiology Report}~\cite{johnson2024mimic}, English chest X‑ray reports from the benchmark database; (2) \textbf{MIMIC Discharge}~\cite{elgaar2024meddec}, English discharge summaries representing semi‑structured patient outcome documents; and (3) \textbf{In‑house Radiology Report}, Chinese abdominal CT reports collected in 2025 from a hospital in China.

We selected 500 samples from each of the three data sources and split each dataset 60:40 into training and testing subsets. Outputs from the training set populated memory for RAG-based models and MedTextWeaver. The testing set is used in the main experiment for model comparison. 

\subsubsection{Medical Text Quality Evaluation}
\label{subsec:experimental_quality_evaluation}

We evaluate four quality dimensions: (1) \textbf{Correctness}: Edited medical text must preserve the accuracy of radiology findings. Using the original report as a reference, correctness is measured by two validated language metrics, SRR-BERT and RadEval-BERT, which detect significant clinical errors in edited reports~\cite{xu2025radeval, delbrouck2025automated}. To address their insensitivity in Chinese radiology reports, we refine a Chinese-ROUGE score with domain-specific medical terms~\cite{chen2026rougechinese}; (2) \textbf{Formatting}: Adherence to report formatting rules is a central goal of medical text editing. We convert established English and Chinese formatting guidelines~\cite{acr2026practice, zj2023standards} into rubric systems, evaluated with LLM-as-a-Judge following the common practice~\cite{arora2025healthbench, liang2026evaluating}; (3) \textbf{Readability}: Following Rooney’s work on medical document literacy~\cite{rooney2021readability}, we apply three conventional readability metrics, Flesch-Kincaid (FK)~\cite{flesch1948new}, Gunning Fog~\cite{gunning1952technique}, and SMOG~\cite{mc1969smog}, to assess public accessibility. 
The details of the above metrics, prompt designs for format compliance, and examples are in the supplementary material.

\subsubsection{Evaluation Metrics}
\label{subsec:experimental_evaluation_metrics}

We define a target profile for medical text editing with three criteria: (1) \textbf{Readability}, constrained to the range of 9th–12th grade, a moderate level for the public; (2) \textbf{Correctness}, requiring at least 85\% agreement with the raw report; and (3) \textbf{Formatting}, requiring at least 75\% compliance with rubric rules. Based on these criteria, we compute pass rates for correctness (PR@C), readability (PR@R), and formatting (PR@F). 

Formally, let $\mathcal{D}$ denote the set of all reports, and $\mathbf{1}_{\{\cdot\}}$ be the indicator function. For a given dimension $X \in \{C,R,F\}$, the pass rate is defined as:

\[
PR@X = \frac{1}{|\mathcal{D}|} \sum_{d \in \mathcal{D}} \mathbf{1}_{\{d \text{ satisfies criterion } X\}}.
\]

The \textbf{Overall Pass Rate (OPR)} measures multi-dimensional compliance, defined as:

\[
OPR = \frac{1}{|\mathcal{D}|} \sum_{d \in \mathcal{D}} \mathbf{1}_{\{d \text{ satisfies } C \land R \land F\}}.
\]

Evaluation metric details are provided in the supplementary material. 

\subsection{Setups of Main experiments}
\label{subsec:experimental_main_setups}
For the main experiments, we adopted \textit{Deepseek-chat}~\cite{deepseek_basic} as the base model and evaluated MedTextWeaver against two categories of baselines. 

First, to assess the impact of the experience weaver, we compared MedTextWeaver against state-of-the-art LLMs, including \textit{Deepseek v4 pro} and \textit{Gemini-3.5 flash}~\cite{deepmindGemini}. These models directly revised raw clinical text without additional modular designs. 

Second, we compared MedTextWeaver with three paradigms of training-free LLM agents: (1) RAG-based approaches, including standard RAG~\cite{lewis2020retrieval}, reflection-oriented RAG, and error-oriented RAG; (2) memory-based approaches, including Agent Workflow Memory~\cite{wang2024agent} and MemoryBank~\cite{zhong2024memorybank}; and (3) evolving-agent approaches, including ExpeL~\cite{zhao2024expel} and EvolveR~\cite{wu2025evolver}. To ensure a fair comparison, all baselines are re-implemented and evaluated with shared reflective observations as the memory source. Agent base models and configurations are strictly identical. 

A complete list of baseline models and configurations is provided in the supplementary material, with details of comparative methods also included.

\subsection{Setups of Real-world Validation Experiments}
\label{subsec:experimental_validation_setups}

In addition to the main simulation experiments, we conducted a validation study involving human‑labeled error detection to ensure real‑world reliability. The dataset comprised 1,000 radiology report errors, annotated in a blinded, head-to-head manner by two medical experts (Dr. A with 10 years of experience and Dr. B with 6 years), which served as the ground truth. Performance was measured using Accuracy, Macro‑Precision, and Macro‑Recall. A case study on error detection further illustrates how MedTextWeaver operates in practice.

\begin{table*}[t]
\centering
\small
\setlength{\tabcolsep}{1mm}

\begin{tabular}{lccccccccc}
\hline
 Models & \multicolumn{3}{c}{DeepSeek-flash~\cite{deepseek_basic}} & \multicolumn{3}{c}{Mistral-Large~\cite{mistralIntroducingMistral}} & \multicolumn{3}{c}{GPT 5.1~\cite{openaiIntroducingGPT5}} \\
\cline{2-10}
 Metric & Accuracy & Precision & Recall & Accuracy & Precision & Recall & Accuracy & Precision & Recall \\
\hline
Baseline & 0.692 & 0.706 & 0.629 & 0.726 & 0.720 & 0.692 & 0.702 & 0.714 & 0.645\\
ExpeL (no rule) & 0.692&0.706&0.630&0.726&0.720&0.692&0.744&0.733&0.672 \\
ExpeL (with rule) &0.762&0.750&0.710& 0.785 & 0.781&0.796 &0.712&0.750 &0.672 \\
EvolveR &0.720&0.753&0.688&0.832&0.817&0.714&0.814&0.716&0.647 \\
\hline
\multicolumn{10}{l}{\textit{Baseline model}} \\
\quad + MedTexWeaver ($\tau$ = 1) & 0.834 & 0.805  & 0.736  & 0.851 & 0.815 & \textbf{0.890} & 0.724 & 0.734 & 0.689 \\
\quad + MedTexWeaver ($\tau$ = 3) & 0.826 & \textbf{0.821}  & \textbf{0.747}  & \textbf{0.874} & \textbf{0.831} & 0.813 & \textbf{0.818} & \textbf{0.794} & \textbf{0.728} \\
\quad + MedTexWeaver ($\tau$ = 5) & \textbf{0.846} &  0.814 & 0.743  & 0.840 & 0.807 & 0.748 & 0.812 & 0.773 & 0.726 \\
Improvement (\%) & +19.5\% & +16.3\% & +18.7\% & +20.4\% & +15.4\% & +28.6\% & +16.5\% & +11.2\% & 12.9\% \\\hline
\end{tabular}%
\caption{Performance of MedTextWeaver on a real-world error detection task. Three moderate LLMs were chosen as backbone models. $\tau$ denotes the maximum number of procedural knowledge items injected. 
}
\label{tab:error_detection}
\end{table*}

\section{Experimental Results}
\label{sec:experimental_results}

\subsection{Comparative Results}
\label{subsec:results_comparative_results}

\paragraph{Overall performance.}  

Table~\ref{tab:main} shows that transforming fragmented evaluations into structured knowledge enables more effective medical text adaptation than directly reusing feedback or relying on general-purpose LLM knowledge. MedTextWeaver consistently improves the quality of medical text across three datasets and multiple evaluation dimensions. Compared with DeepSeek-v4-flash without acquired knowledge, MedTextWeaver improves the Overall Pass Rate by 34.67\%, 4.40\%, and 17.00\% on MIMIC Chest X-ray reports, MIMIC Discharge summaries, and the in-house Abdominal CT dataset, respectively. We attribute these gains primarily to the integration of acquired principles and procedural knowledge, which equips the framework with a more intrinsic, context-aware understanding of multi-faceted medical text quality criteria.

Beyond simple knowledge injection, we further compared our method against representative memory-based approaches. MedTextWeaver consistently exhibits superior performance across all three quality dimensions, with a particularly notable advantage in the preservation of fine-grained clinical details. A detailed discussion of this phenomenon, along with qualitative sample comparisons, is deferred to the ablation study and supplementary materials. 

Moreover, unlike frontier general-purpose models that tend to over-optimize a single metric (typically factual correctness) at the cost of other quality dimensions, our approach effectively mitigates this multi-dimension conflict by fusing procedural guidance linked to particular types of errors. These results collectively validate the efficacy of injecting procedural knowledge into editing various medical texts, and we next conduct a real-world validation experiment and ablation study.


\begin{figure}[t]
  \centering
  \includegraphics[width=\columnwidth]{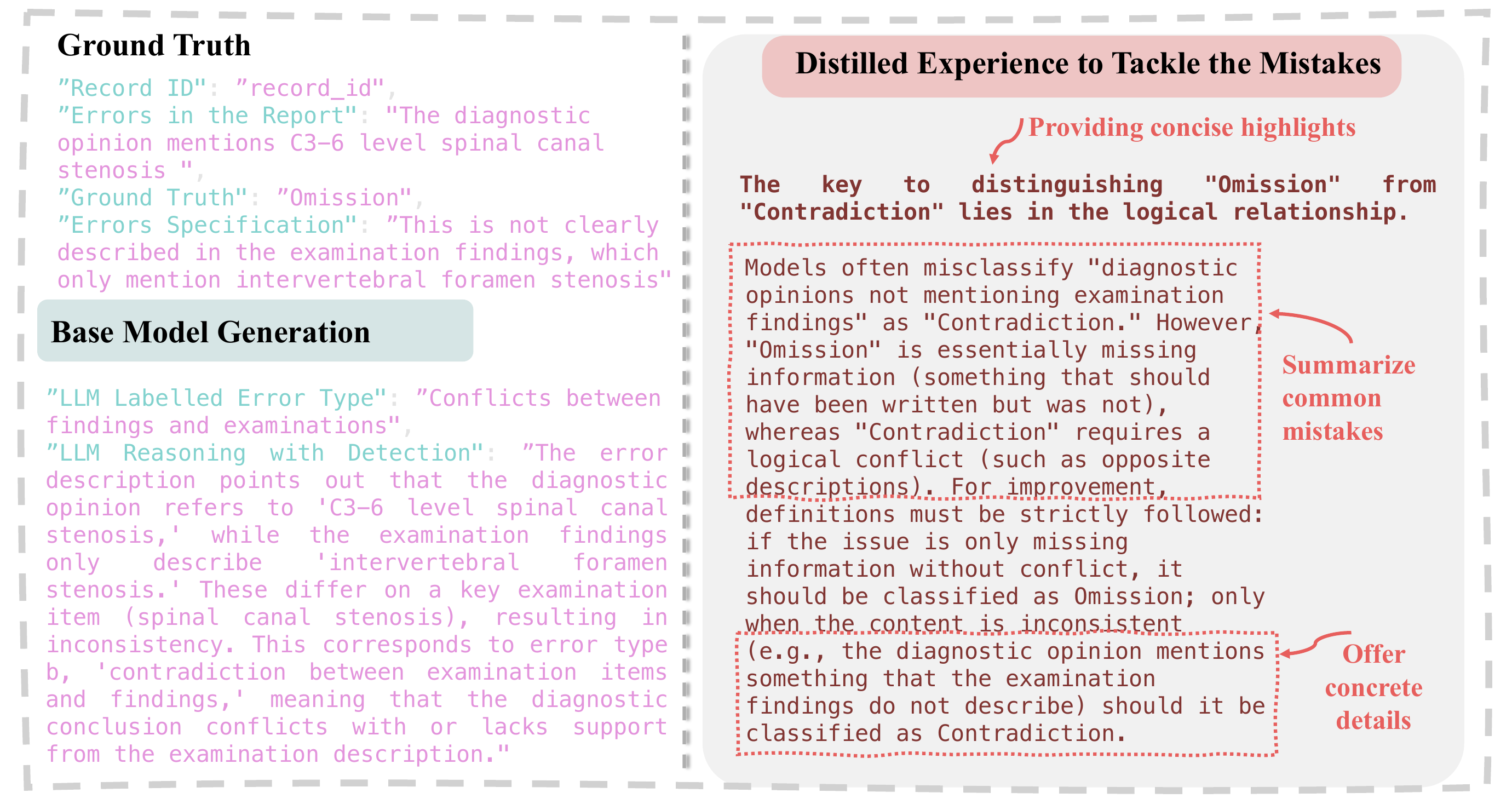}
  \caption{Case Study. Explanation of how experience improves detection accuracy.}
  \label{fig:case_study}
\end{figure}

\subsection{Real-world Validation Experiments of Error Detection}
\label{subsec:results_error_detection}

Table~\ref{tab:error_detection} evaluates whether synthesized knowledge improves the agent’s ability to identify clinically relevant errors. Across three base models, MedTextWeaver consistently improves error detection performance, increasing accuracy, precision, and recall by up to 20.4\%, 16.3\%, and 28.6\%, respectively. These results demonstrate the effectiveness of MedTextWeaver in medical text editing under limited expert supervision, particularly in scenarios where subtle misalignments exist between clinicians’ definitions of errors and the intrinsic knowledge encoded in LLMs. 

MedTextWeaver mitigates these misalignments by transforming fine-grained reflections on individual errors into actionable, high-level knowledge representations. The case study in Figure~\ref{fig:case_study} further illustrates the multi-layered knowledge distilled by MedTextWeaver, which explains the consistent performance gains observed in this experiment and in Table~\ref{tab:main}.

\section{Ablation}
\label{sec:ablation}

To evaluate the contribution of individual components in MedTextWeaver, we conducted ablation studies using Deepseek-V4 Flash as the backbone. We compared the full model against five variants: (1) w/o Hierarchical Distillation: removing tree-based knowledge weaving and directly passing reflections to later stages; (2) w/o Cross-dimension Fusion: eliminating fusion across dimensions and proceeding separately to procedural distillation; (3) w/o Procedural Knowledge Conversion: using distilled knowledge directly for strategy condensation without contextual tips; (4) w/o Principles: excluding strategy input and retaining only procedural knowledge; and (5) \textbf{w/o Concreteness Preservation}: removing emphasis on concrete detail preservation in prompts. Table~\ref{tab:ablation} summarizes the performance impact of each module, showing that all components contribute to the overall efficacy of the agent.

\begin{table}[t]
\centering
\small
\setlength{\tabcolsep}{1mm}
\begin{tabular}{@{}lcccc@{}}
\hline
Method & PR@C & PR@R & PR@F & OPR \\
\hline
MedTextWeaver & 68.00 & 66.00 & 38.00 & 19.00\\
\quad w/o Hierarchical Distillation & 67.00 & 61.00 & 34.00 & 14.00 \\
\quad w/o Cross-dimension Fusion & 71.00 & 62.00 & 34.00 & 20.00 \\
\quad w/o Procedural Knowledge & 43.88 & 67.35 & 1.02 & 0.00 \\
\quad w/o Principles & 73.00 & 65.00 & 37.00 & 17.00 \\
\quad w/o Concreteness Preservation & 43.00 & 39.00 & 14.00 & 3.00 \\
\quad Original Report & 100.0 & 59.00 & 0.00 & 0.00 \\
\hline
\end{tabular}
\caption{Ablation study of MedTextWeaver components on MIMIC-IV Chest X-ray reports.}
\label{tab:ablation}
\end{table}

\paragraph{Effects of Hierarchical Distillation with Detail Preservation.} 
Removing hierarchical distillation and cross-dimensional fusion affects weaker dimensions more, with readability and formatting compliance decreasing by 5\% and 4\%, respectively. These effects are partially mitigated in later stages, such as procedural knowledge and high-level principles, since information passes through these modules before being injected into the context. Consequently, eliminating these preliminary stages primarily reduces the efficacy of subsequent modules, particularly in weaker-performing dimensions where the evaluation standards deviate more strongly from the intrinsic knowledge of LLMs.

\paragraph{Roles of High-level Principles and Fine-grained Procedural Knowledge.} 
Quality dimensions exhibit distinct sensitivities to high-level versus fine-grained information. Removing procedural knowledge leads to substantial reductions in correctness and formatting compliance, while removing principles impacts more on readability. This pattern aligns with the intuition that correctness and formatting require fine-grained regulatory detail, whereas readability adjustments depend more on general audience perception. Together, high-level principles and procedural knowledge play complementary roles in improving medical text quality across all dimensions.

\paragraph{Necessity of Concreteness Preservation.} 
Concreteness preservation plays a critical role in ensuring effective synthesis of hierarchical distillation and procedural knowledge. Removing the emphasis on concreteness in prompts results in significant performance degradation across all three quality dimensions. This distinguishes our framework from conventional memory summarization-oriented approaches, demonstrating that actionable, detail-preserving procedural knowledge is essential for medical text editing.

\section{Related Work}
\label{sec:related_work}

\subsection{LLM-empowered Medical Text Editing}
\label{subsec:related_llm_medical}

LLMs have been increasingly applied to medical text processing, with existing studies focusing primarily on structured reporting~\cite{bergomi2024reshaping, sacoransky2024chatgpt}, translation~\cite{guerreiro2024xcomet}, and information extraction~\cite{le2024performance}. Recent efforts have further explored knowledge editing to improve factuality and reduce hallucination in medical contexts~\cite{xu2024editing, chen2026beyond}. However, improving medical text editing remains challenging when supervision is sparse, fragmented, and distributed across multiple quality criteria. This work investigates how heterogeneous expert evaluations can be transformed into reusable knowledge for adapting LLM agents under limited supervision.

\subsection{Self-evolving Agent}
\label{subsec:related_self_evolving_agent}

Self-evolving agents have emerged as an important direction for improving LLM capabilities when large-scale training signals are unavailable. Existing approaches emphasize experience accumulation through reflection, critique, summarization, and memory management mechanisms~\cite{wang2024agent, zhao2024expel, qu2024exploration, zhang2025adaptive, zheng2025skillweaver}. However, accumulating individual experiences does not necessarily produce a coherent understanding of complex tasks with interacting objectives. Our work complements existing self-evolving agents by studying how fragmented feedback can be progressively transformed into structured knowledge that captures both general principles and actionable behaviors.



\section{Conclusion}
\label{sec:conclusion}

Beyond medical text editing, our study highlights a fundamental limitation of current feedback-driven adaptation: observations alone do not constitute knowledge unless an agent can discover the underlying principles that connect them. By transforming fragmented evaluations into structured and reusable representations, MedTextWeaver provides a framework for studying how LLM agents can evolve their task-specific understanding. This perspective opens a direction toward more reliable and interpretable agent adaptation in domains where expert supervision is valuable but inherently sparse and heterogeneous.


\bibliography{ref}

\appendix

\section{Overview of the Supplementary Material}
\label{sec:appendix}

This supplementary material provides additional details to support the main paper. It includes further experiment results (Appendix~\ref{app:further_experiments}), evaluation metric details (Appendix~\ref{app:evaluation_details}), baseline model configurations (Appendix~\ref{app:llm_baselines}), details of comparative methods (Appendix~\ref{app:details_comparative_method}), representative knowledge-weaving examples (Appendix~\ref{app:what_knowledge}), and prompt designs and examples (Appendix~\ref{app:prompts}).

\section{Further Experiment Results}
\label{app:further_experiments}

Table~\ref{tab:score_diff} reports percentage score changes between original and revised reports across the three clinical text datasets, using the same evaluation dimensions as the main experiments. By design, Diff@C is always negative; smaller absolute values indicate less loss of key medical information during revision. Readability differences are harder to interpret because a score change may either remain within the expected range or diverge from it. For formatting, where scores increase monotonically with compliance, MedTextWeaver outperforms both baseline models and memory-based methods.

\begin{table*}[t]
\centering
\small
\setlength{\tabcolsep}{1mm}
\begin{tabular}{lccccccccc}
\toprule
Data Source & \multicolumn{3}{c}{MIMIC Chest X-ray} & \multicolumn{3}{c}{MIMIC Discharge} & \multicolumn{3}{c}{In-house Abdominal CT}  \\
\cmidrule{2-10}
Score Difference & Diff@C & Diff@R & Diff@F & Diff@C & Diff@R & Diff@F & Diff@C & Diff@R & Diff@F \\
\midrule
\multicolumn{10}{l}{\textbf{Section 1: SOTA LLMs Baseline}} \\
\quad Deepseek-v4-pro & -8.91\% & -2.05\% & +5.67\% & -9.73\% & +4.13\% & +2.87\% & -2.61\% & -0.15\% & +2.98\% \\
\quad Gemini-3.5-flash & -10.10\% & -2.84\% & +7.00\% & -6.56\% & -0.35\% & +3.77\% & -5.65\% & +0.06\% & +5.38\% \\
\midrule
\multicolumn{10}{l}{\textbf{Section 2: Comparative RAG/Memory Methods}} \\
\quad RAG (Detection) & -9.19\% & -3.25\% & +9.69\% & -4.15\% & -1.09\% & +7.29\% & -7.28\% & -0.50\% & +0.80\% \\
\quad RAG (Revision) & -9.66\% & -1.86\% & +8.56\% & -4.69\% & +0.15\% & +4.67\% & -5.23\% & -0.29\% & -0.66\% \\
\quad Agent Workflow Memory & -8.25\% & -3.32\% & +5.26\% & -7.22\% & +1.13\% & +5.24\% & -3.56\% & -0.36\% & +1.49\% \\
\quad MemoryBank & -8.30\% & -2.44\% & +47.23\% & -7.07\% & +6.05\% & +8.51\% & -3.59\% & -0.31\% & +3.83\% \\
\quad ExpeL & -6.92\% & -0.72\% & +3.85\% & -6.76\% & +0.36\% & +4.65\% & -3.69\% & -0.26\% & +0.09\% \\
\quad EvolveR & -7.71\% & -1.40\% & +12.33\% & -7.13\% & -0.12\% & +7.40\% & -3.80\% & -0.36\% & +3.55\% \\
\midrule
\multicolumn{10}{l}{\textbf{Section 3: Weaving Experience}} \\
\quad Deepseek-v4-flash & -11.48\% & -2.84\% & +9.85\% & -6.94\% & +12.67\% & +7.61\% & -3.14\% & -0.32\% & +4.72\% \\
\quad + MedTexWeaver(Detection) & -4.41\% & -2.76\% & +44.07\% & -2.40\% & -3.21\% & +2.21\% & -6.06\% & +0.56\% & +4.68\% \\
\quad + MedTexWeaver(Revision) & -7.97\% & -2.22\% & +36.41\% & -3.04\% & -2.74\% & +8.07\% & -3.49\% & -0.29\% & +4.04\% \\
\quad + MedTexWeaver(Full) & -4.79\% & -1.68\% & +43.45\% & -2.86\% & -1.36\% & +7.97\% & -6.22\% & +0.62\% & +8.47\% \\
\bottomrule
\end{tabular}
\caption{
Percentage score change across data sources and evaluation metrics. Diff@C = score difference for correctness, Diff@R = score difference for readability, and Diff@F = score difference for format.
}
\label{tab:score_diff}
\end{table*}

\section{Evaluation Metric Details}
\label{app:evaluation_details}

\subsection{Correctness}
Ensuring consistency between the raw and revised medical text is critical. SRR-BERT and RadEval-BERT are two metrics originally designed to evaluate automatically generated radiology reports against ground truth~\cite{xu2025radeval, delbrouck2025automated}. These metrics assess the correct presence of medical terms and address biases inherent in conventional ROUGE and BERTScore, particularly when dealing with semantically similar terminology. Accordingly, we adopted SRR-BERT and RadEval-BERT to evaluate whether revised reports preserve clinically important information. However, both metrics exhibit reduced sensitivity in Chinese radiology reports. To mitigate this limitation, we refined a Chinese-ROUGE score with domain-specific medical terminology~\cite{chen2026rougechinese}.

\subsection{Readability}
We employed three conventional readability metrics, Flesch-Kincaid (FK)~\cite{flesch1948new}, Gunning Fog~\cite{gunning1952technique}, and SMOG~\cite{mc1969smog}, to assess public accessibility. Each metric estimates the grade level required for comprehension. Scores above grade 15 indicate texts demanding professional expertise, approaching the difficulty of academic publications and potentially hindering patient communication. By contrast, setting the target readability to grades 9–12 corresponds to a high school level, which is generally accessible to the broader public.

\subsection{Rubrics}
Adherence to report formatting rules is a central objective in medical text editing. Currently, no established automated tool exists to assess compliance with formatting standards. In practice, clinicians rely on radiology and discharge report guidelines to improve writing quality. Following this principle, we converted established English and Chinese formatting standards~\cite{acr2026practice, zj2023standards} into rubric systems, which are evaluated using LLM-as-a-Judge in line with common practice~\cite{arora2025healthbench, liang2026evaluating}. The rubrics are aligned with the sections available in the dataset to avoid mismatches between the rubric requirements and the content of the report. The full rubric definitions are organized into three categories below.

\begin{lstlisting}
"demographics": {
  "facility": "The report states the facility or location where the study was performed.",
  "patient_identity": "The report includes the patient's name and date of birth or age.",
  "patient_sex_or_gender": "The report includes the patient's sex assigned at birth or gender.",
  "referring_provider": "The report names the referring physician or states the patient is self-referred.",
  "examination_name": "The report clearly states the name or type of imaging examination performed.",
  "examination_date": "The report includes the date on which the imaging examination was performed.",
  "clinical_information": "The report includes relevant clinical information supporting the indication for the study."
}
\end{lstlisting}

\begin{lstlisting}
"body_of_the_report": {
  "procedures_and_materials": "The report describes the procedures performed and any contrast media, radiopharmaceuticals, medications, or devices used.",
  "findings_terminology": "The report describes imaging findings using appropriate anatomic, pathologic, and radiologic terminology.",
  "limitations": "The report identifies factors that may limit the sensitivity or specificity of the examination.",
  "clinical_question": "The report addresses the specific clinical question, or explains why it cannot be answered.",
  "comparison_studies": "The report compares findings with relevant prior examinations and reports when available and appropriate."
}
\end{lstlisting}

\begin{lstlisting}
"impression": {
  "separate_section": "The report contains a separate impression or conclusion section unless the report is brief.",
  "specific_diagnosis": "The impression provides a specific diagnosis when the findings allow one.",
  "differential_diagnosis": "The impression provides a differential diagnosis when a single diagnosis is not possible.",
  "follow_up_recommendation": "The impression suggests appropriate follow-up or additional diagnostic studies when needed.",
  "adverse_event": "The impression briefly notes any significant adverse event related to the study."
}
\end{lstlisting}

\section{LLM Baselines and Configurations}
\label{app:llm_baselines}

Our baseline models were selected based on state-of-the-art LLMs available at the time of experimentation (July 2026).

Domain-specific fine-tuned medical LLMs were not included, as general-purpose state-of-the-art models have consistently demonstrated superior performance in the medical domain. For example, \citet{kim2025questioning} reported that GPT-4 achieved the highest performance across multiple medical benchmarks, outperforming both human doctors and domain-specialized models such as Meditron-7B and BioMistral-7B. This trend, in which general-purpose LLMs surpass medically fine-tuned variants, has only strengthened with newer models.

All LLM baselines were configured identically unless otherwise noted: the sampling temperature was set to 0.7 and the maximum output length was capped at 2048 tokens. For the two DeepSeek models (Deepseek-v4-pro and Deepseek-v4-flash), the reasoning effort was set to high. These settings were applied consistently across the three baselines---Gemini 3.5-flash, Deepseek-v4-pro, and Deepseek-v4-flash---and to the LLM components of the comparative methods.

\section{Details of Comparative Methods}
\label{app:details_comparative_method}
\subsection{Reproduction details}

\begin{itemize}
    \item \textbf{Agent Workflow Memory.} AWM induces reusable revision workflows from a collection of original/revised report pairs and their reflective feedback. An LLM reads batches of experiences and induces concise, reusable revision workflows. At inference time, the current report is embedded and the most relevant workflows are retrieved by dense embedding similarity using pure NumPy. A generated workflow is a common editing subroutine such as "replace vague quantifiers with concrete measurements" or "expand abbreviations to full clinical terms".
    \item \textbf{MemoryBank.} The adapter borrows MemoryBank's hierarchical memory structure, spanning session memories, overall principles, and a profile. It is realized with an LLM that summarizes each session into a concise memory, aggregates all session summaries into long-term, overarching guidance, and finally extracts a profile of common error tendencies and recommended revision strategies. At inference time, the query is embedded and the most relevant session memories are retrieved, together with the overall memory and profile.
    \item \textbf{ExpeL.} Instead of importing the original ExpeL codebase, the adapter borrows its core concept: an LLM reads batches of revision experiences and iteratively refines a small set of general, transferable rules/insights. For rule generation, an LLM reads batches of experiences and iteratively refines a short list of general, transferable rules/insights using ExpeL-style AGREE / REMOVE / EDIT / ADD operations. At inference time, the current report is embedded and the most relevant rules are retrieved by dense embedding similarity using pure NumPy.
    \item \textbf{EvolveR.} EvolveR distills guiding and cautionary principles from a collection of revision experiences. An LLM reads batches of experiences and distills guiding and cautionary principles as structured JSON. At inference time, the current report is embedded and the most relevant principles are retrieved by dense embedding similarity using pure NumPy. A generated principle is a concise, generalizable piece of advice such as "Use full clinical terms instead of ambiguous abbreviations" or "Do not combine multiple distinct findings into a single impression sentence".
\end{itemize}

All RAG/Memory methods used the same source files as raw memory sources and were run with Deepseek-v4-flash using the same configuration.

\subsection{Memory Comparison between Methods}

The following subsections illustrate how different methods retain and reuse information during medical text editing. RAG-based methods retrieve external style-guide passages, memory-based methods store raw historical examples or distilled rules, and MedTextWeaver accumulates procedural knowledge. Comparing memory-based approaches, MemoryBank and MedTextWeaver generate longer instructions than Agent Workflow Memory, ExpeL, and EvolveR. This observation is consistent with their superior performance in the compliance with the formatting. It also aligns with the findings of the ablation study, which highlight the importance of preserving concrete details in accumulated experience. Both MemoryBank and MedTextWeaver demonstrate better performance in detail-sensitive metrics such as formatting correction. Furthermore, compared with MemoryBank, MedTextWeaver provides richer detail in professional medical terminology, which enhances the preservation of medical integrity after revision. As a result, MedTextWeaver achieves higher correctness and overall performance than MemoryBank.

\subsubsection{Agent Workflow Memory}
Here is a memory sample generated by Agent Workflow Memory.

\begin{lstlisting}[basicstyle=\footnotesize\ttfamily,breaklines=true,breakatwhitespace=true]
{"workflow_index": 0, "workflow_name": "expand_clinical_abbreviations", "workflow_body": "When a medical report contains non-standard or ambiguous abbreviations in the indication or findings, expand them to their full clinical terms.
1. Identify abbreviations that may be unclear to the intended reader (e.g., 'PTX', 'PNA', 'r/o', 'SBO', 's/p', 'CHF').
2. Replace each abbreviation with its full term while preserving the original clinical meaning.
3. Ensure the surrounding punctuation and grammar remain natural."}
\end{lstlisting}

\subsubsection{MemoryBank}
Here is a memory sample generated by MemoryBank.

\begin{lstlisting}[basicstyle=\footnotesize\ttfamily,breaklines=true,breakatwhitespace=true]
{"entry_id": "memorybank_session_0", "role": "memory_bank_session", "content": "### Core Revision Patterns

- **Placeholder degradation**: Underscore placeholders (`___`) for missing dates or data are often mistakenly replaced with bracketed words (e.g., `[date]`) or removed entirely, violating standard radiology format conventions.
- **Abbreviation expansion without benefit**: Common clinical shorthand (`PTX`, `PNA`, `r/o`, `SBO`) is sometimes spelled out, needlessly increasing syllable count and text complexity without improving clarity for the intended audience.
- **Impression restructuring**: Separate impression statements are frequently merged into a single compound sentence, reducing structural clarity. In other cases, findings from the body are added to the impression, altering the originally documented conclusion. Critical descriptors (e.g., "innumerable") may be dropped.
- **Terminology alteration**: Precise medical descriptions are swapped for simpler or different terms (e.g., "moderately dense fibroglandular and fibronodular" -> "scattered fibroglandular densities"), which changes clinical specificity and may alter the report's meaning.
- **Missing structural elements**: Reports often lack required sections such as a patient header, a separate Impression section, or a radiologist signature, violating standard format requirements.

### Common Issues

- **Incorrect placeholder handling** - using `[date]` or generic words instead of underscores, or deleting placeholders altogether.
- **Unintended meaning shifts** - removing blank fields, dropping qualifiers, or narrowing/expanding descriptive scope (e.g., "concerning osseous lesion" -> "suspicious or aggressive osseous lesion").
- **Impression integrity loss** - merging items, importing body findings, or simplifying key clinical terms.
- **Inconsistent internal references** - mixing "prior" and "previous" within the same paragraph.
- **Format noncompliance** - no patient header, missing Impression section, no reporting radiologist identifier.

### Effective Revision Strategies

- **Retain placeholders as underscores (`___`)** for all missing data; do not substitute with words or brackets.
- **Preserve standard abbreviations** unless the reading level specifically requires expansion, and only then if complexity does not significantly increase.
- **Keep impression content exactly as stated** - do not merge statements, add new findings, or delete clinically"}
\end{lstlisting}

\subsubsection{ExpeL}
Here is a memory sample generated by ExpeL.

\begin{lstlisting}[basicstyle=\footnotesize\ttfamily,breaklines=true,breakatwhitespace=true]
{"rule_index": 0, "rule": "Preserve all factual clinical observations and descriptors; do not substitute synonyms that may alter clinical meaning unless corrected by verified source data"}
{"rule_index": 1, "rule": "Retain standard medical abbreviations and short forms that enhance conciseness and readability for the intended professional audience"}
\end{lstlisting}

\subsubsection{EvolveR}
Here is a memory sample generated by EvolveR.

\begin{lstlisting}[basicstyle=\footnotesize\ttfamily,breaklines=true,breakatwhitespace=true]
{"principle_index": 0, "type": "guiding", "description": "Use underscores as placeholders for missing dates or values instead of bracketed text like '[date]'.", "structure": [["placeholder", "should_be", "underscore"], ["bracketed_text", "is_not", "standard_placeholder"]]}
{"principle_index": 1, "type": "cautionary", "description": "Do not replace standard underscore placeholders with bracketed text, as this violates formatting conventions.", "structure": []}
\end{lstlisting}

\subsubsection{MedTextWeaver}
Here is a memory sample generated by MedTextWeaver.

\begin{lstlisting}[basicstyle=\footnotesize\ttfamily,breaklines=true,breakatwhitespace=true]
1. Verify medication entry formatting: compare each medication line to the original, ensuring delimiters such as colons between frequency and PRN reason (e.g., 'Q8H:PRN pain') and asterisks around Disp and Refills (e.g., 'Disp:*30 Tablet Refills:*0') are preserved, and that RX details remain on separate lines rather than merged into sentences.
2. Check placeholder elements: scan for bracketed placeholders like '[dose]' or '[Patient Name]' and ensure they match the source's use of underscores '__'; also confirm all required structural sections (patient header, admission diagnosis, facility name) are present with either actual values or the original blank underscores, never replaced with assumptions or alternative notations.
3. Detect missing discharge date: search the full text for a line containing 'Discharge Date' or 'Date of Discharge' followed by a date; if absent, this critical temporal marker must be inserted in the correct position (after the header or before Discharge Diagnosis) to maintain report completeness.
4. Validate structural completeness: confirm that all mandatory sections---such as patient demographics, admission and discharge diagnoses, medications, disposition, and follow-up instructions---are included and properly ordered; missing or misplaced sections are common causes of low-quality discharge documentation.
5. Enforce format consistency: review the revision for any rephrasing or restructuring that deviates from the source template, such as converting structured lists into prose or altering standard label patterns; faithful reproduction of the original format is essential for clinical accuracy and readability.

Procedural Knowledge:
### missing discharge date

Search the full text for any line containing 'Discharge Date' or 'Date of Discharge' followed by a date. If absent, insert 'Date of Discharge: [date]' after the header or before Discharge Diagnosis.

Verify the presence of a 'date of discharge' label using regex; if missing, add 'Discharge Date: [DD/MM/YYYY]' after the Discharge Disposition section or before medications to ensure structural completeness.
\end{lstlisting}

\section{What Knowledge Is Being Woven?}
\label{app:what_knowledge}

\begin{figure*}[htbp]
  \centering
  \includegraphics[width=\textwidth]{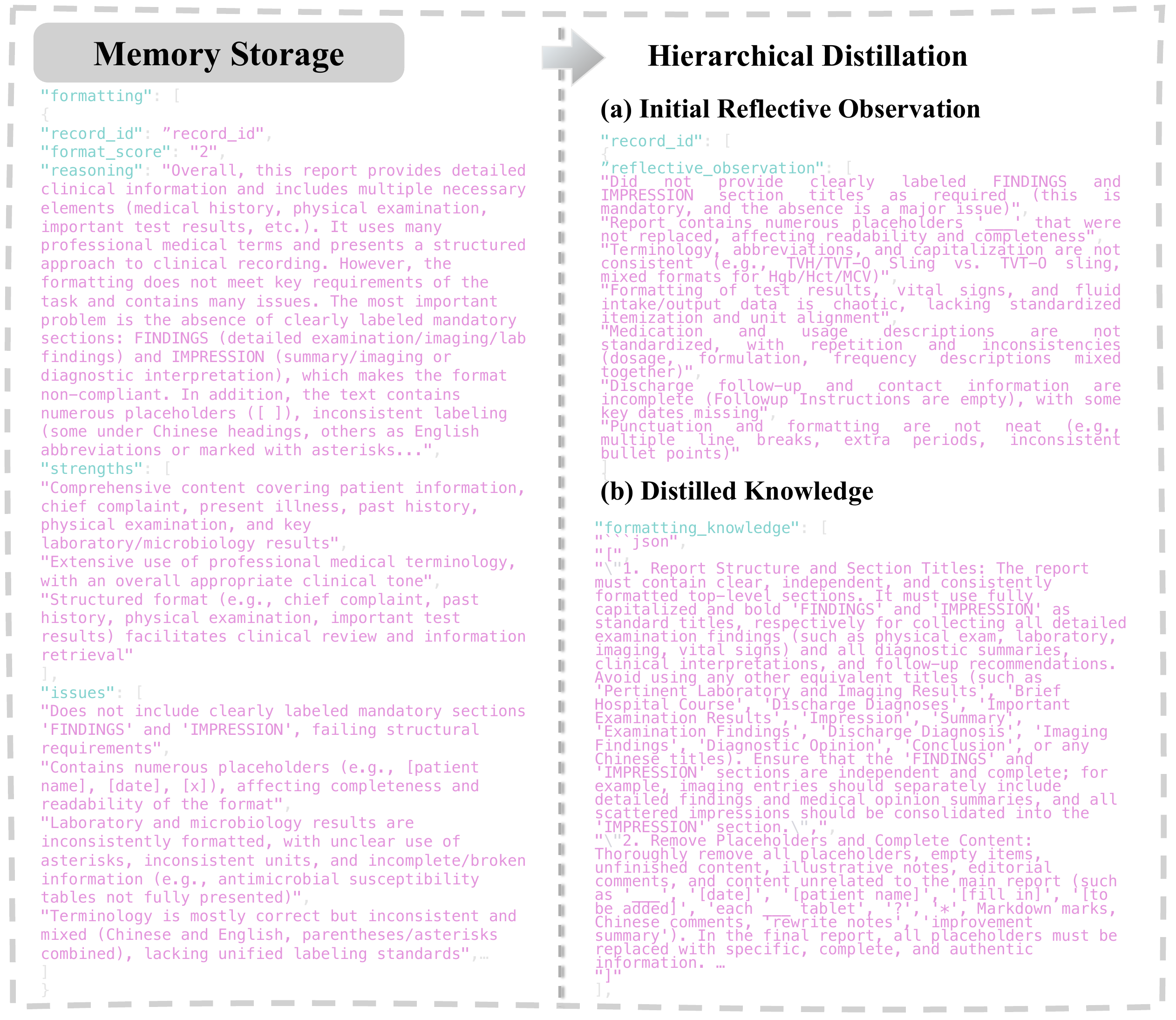}

  \includegraphics[width=\textwidth]{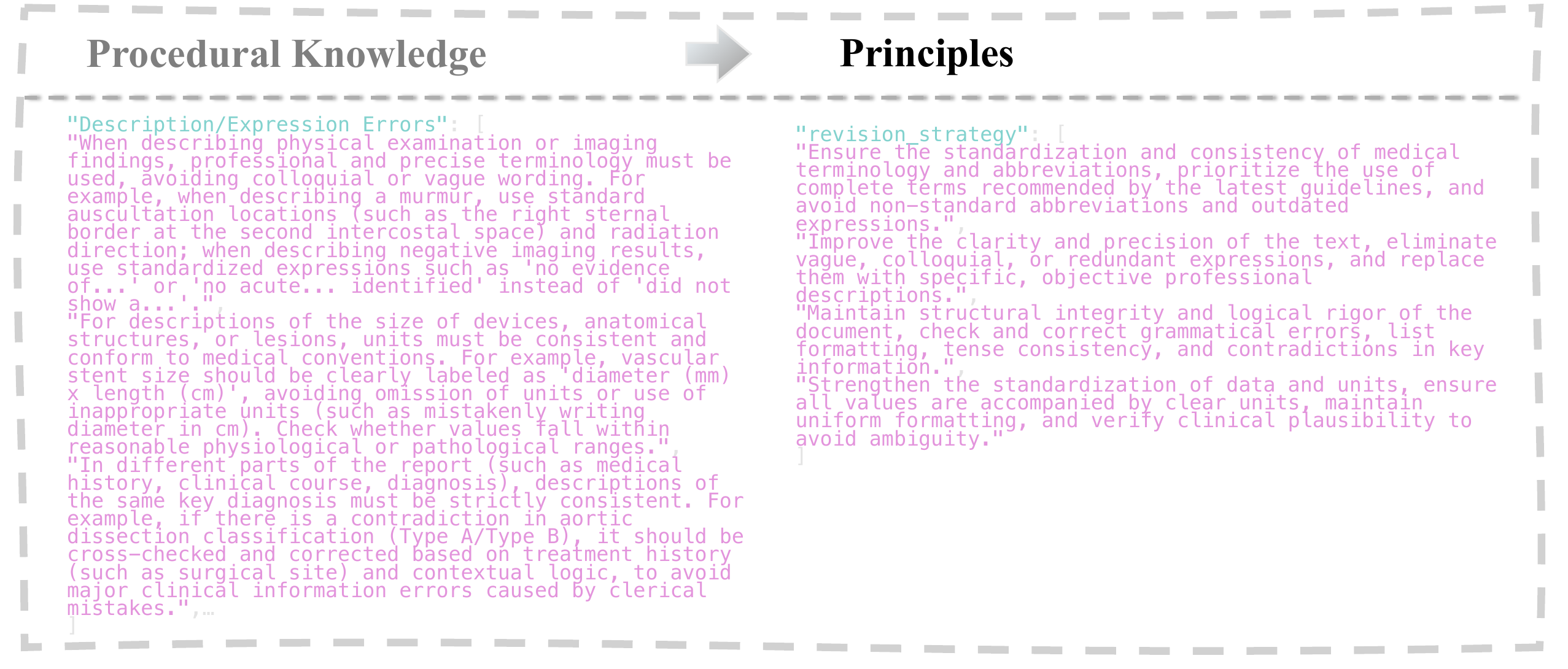}
  \caption{Representative outputs from the MedTextWeaver process. Top: raw feedback is abstracted and distilled into reflective observation. Bottom: the transformation from procedural knowledge to principles.}
  \label{fig:figure_app_what_woven}
\end{figure*}

Figure~\ref{fig:figure_app_what_woven} illustrates representative outputs from the MedTextWeaver process. The framework progressively transforms raw reflections into structured knowledge through multiple layers. At the initial stage, point-wise reflective observations capture localized errors. These are subsequently consolidated into reflective knowledge that integrates both mainstream patterns and fine-grained details. Building on this foundation, procedural knowledge emerges as actionable strategies for error correction. Finally, high-level principles are distilled to provide generalizable guidance to the agent. This hierarchical progression bridges the gap between local error identification and global understanding, enabling coherent adaptation across diverse quality dimensions.

\section{Prompt Designs and Examples}
\label{app:prompts}

This section provides the prompt templates used in the MedTextWeaver pipeline. Each prompt corresponds to one stage of procedural-knowledge evolution: reflective observation, hierarchical distillation, cross-dimensional synthesis, error detection, and text revision.

\subsection{Reflective Observation}

The reflective-observation prompt abstracts raw feedback into a structured diagnosis of what went wrong in a revision.

\begin{lstlisting}
You are reviewing a medical text edit. An original report and its
revised version are shown below, along with evaluator feedback.

Original report:
{original_text}

Revised report:
{revised_text}

Evaluator feedback:
{feedback}

Write a concise reflection of roughly 100 words that covers:
1. Why this revision received this score in {label}, with concrete details.
2. What concrete changes could satisfy the improvement goal and relevant rubrics or metrics. Respond with only the reflection paragraph.
\end{lstlisting}

\subsection{Hierarchical Distillation}

The hierarchical-distillation prompt turns a set of observations into reusable procedural knowledge.

\begin{lstlisting}
You are an expert medical editor. Read the following structured revision observations about {text_type} reports in {language} and produce one response that groups similar revision suggestions into shared error categories and explains how to improve text quality for each category.

- Combine observations that belong to the same or closely related error_type.
- For each category, preserve concrete examples and actionable rules from the observations so the guidance is specific enough.

Output ONLY a JSON object:
{"combined_experience": "Your combined revision guidance as one response. You may use bullet points or numbered paragraphs for each error category."}

Observations:
{reflections}
\end{lstlisting}

\subsection{Cross-Dimensional Synthesis}

The cross-dimensional-synthesis prompt resolves conflicts between competing editing objectives (e.g., completeness vs. conciseness).

\begin{lstlisting}
You are an expert medical editor improving how a model handles a common error type in {text_type} reports in {language}.

You are given:
- Error type: {error_type}
- Explanations observed for this error: {error_reasons}

Task: Propose at most {suggestions_count} concise, ranked suggestions for how to best detect and revise this error type. The most important suggestion should come first. Each suggestion should be one integral paragraph, and should preserve concrete details and examples where possible.

Output ONLY a JSON array:
["first suggestion as one integral paragraph", "second suggestion as one integral paragraph", "third suggestion as one integral paragraph"]

A pool of cross-dimensional revision experience:
{fused_pool}

\end{lstlisting}

\subsection{Error Detection}

The error-detection prompt applies accumulated knowledge to flag weaknesses in a new report.

\begin{lstlisting}
You are a medical-report quality checker. Review the report below
and identify any errors or weaknesses using the stored guidelines.

Report:
{report_text}

<Injected Memory/Knowledge if any>

Please output a JSON list. Each item must be an object with exactly these keys:
{"error_detected": "the exact problematic text segment from the report",
"error_reason": "a concise description of the error type and how to revise it"}

If the report has no errors, please output an empty JSON list: []
\end{lstlisting}

\subsection{Text Revision}

The text-revision prompt generates the final edited report conditioned on the detected errors and applicable tips.

\begin{lstlisting}
You are an expert medical text editor. Rewrite the report below to
fix the identified issues while preserving all clinically relevant
information.

<Injected Memory/Knowledge if any>

Original report:
{report_text}

Here are the identified errors and the reasons they need revision:
{detected_errors}

Please output a JSON object with exactly these keys:
{"revised_report": "the full revised report text",
"revision_reason": "a concise summary of the revisions made"}
Do not include any text outside the JSON object.
\end{lstlisting}

\end{document}